\documentclass[10pt,twocolumn,letterpaper]{article}

\usepackage{iccv}   
\usepackage{multirow}

\definecolor{iccvblue}{rgb}{0.21,0.49,0.74}
\usepackage[pagebackref,breaklinks,colorlinks,allcolors=iccvblue]{hyperref}

\usepackage[accsupp]{axessibility}
\def\paperID{60} 
\def\confName{ICCV}
\def\confYear{2025}

\title{Breast Cancer VLMs: Clinically Practical Vision-Language Train-Inference Models}

\author{
Shunjie-Fabian Zheng\textsuperscript{1,2} \quad
Hyeonjun Lee\textsuperscript{2} \quad
Thijs Kooi\textsuperscript{2} \quad
Ali Diba\textsuperscript{2} \\
\textsuperscript{1}Department of Medicine I, LMU University Hospital, LMU Munich, Germany\\
\textsuperscript{2}Lunit Inc.\\
{\footnotesize\texttt{shunjiefabian.zheng@med.uni-muenchen.de, \{hyeonjun1882, tkooi, ali\}@lunit.io}}
}

\begin{document}
\maketitle
\begin{abstract}
Breast cancer remains the most commonly diagnosed malignancy among women in the developed world. Early detection through mammography screening plays a pivotal role in reducing mortality rates. While computer-aided diagnosis (CAD) systems have shown promise in assisting radiologists, existing approaches face critical limitations in clinical deployment - particularly in handling the nuanced interpretation of multi-modal data and feasibility due to the requirement of prior clinical history. This study introduces a novel framework that synergistically combines visual features from 2D mammograms with structured textual descriptors derived from easily accessible clinical metadata and synthesized radiological reports through innovative tokenization modules. Our proposed methods in this study demonstrate that strategic integration of convolutional neural networks (ConvNets) with language representations achieves superior performance to vision transformer-based models while handling high-resolution images and enabling practical deployment across diverse populations. By evaluating it on multi-national cohort screening mammograms, our multi-modal approach achieves superior performance in cancer detection and calcification identification compared to unimodal baselines, with particular improvements. The proposed method establishes a new paradigm for developing clinically viable VLM-based CAD systems that effectively leverage imaging data and contextual patient information through effective fusion mechanisms.
\end{abstract} 
\section{Introduction}
Mammography remains the cornerstone of breast cancer screening programs, with population-based initiatives demonstrating 20-35\% mortality reduction via early detection \cite{elmore2005screening}. However, interpreting screening mammograms presents significant challenges due to the subtle appearance of early malignancies, wide variations in breast parenchymal patterns, and the cognitive burden of reviewing hundreds of studies daily \cite{evans2013if}. The CAD models are developed with various complexity to help with malignancy classification and detecting subtle findings like calcification\cite{lehman2015diagnostic}. Current CAD models, while performing greatly for microcalcifications and masses (75-89\%) on public datasets (VinDr\cite{vindr}), still have noticeable performance gaps in real-world clinical data with diverse screening population and more variant malignant cases.


Recent advances in multi-modal deep learning have sought to address these limitations by integrating complementary data sources. Mammo-CLIP~\cite{Ghosh2024MammoCLIP} demonstrated that contrastive language-image pre-training improves malignancy detection accuracy by aligning multi-view mammograms with radiological reports. Similarly, MMBCD~\cite{jain2024mmbcd} introduced breast region-of-interest detection with multi-instance learning to handle high-resolution $2K \times 2K$ mammograms, achieving superior F1-scores by fusing Vision Transformer (ViT) visual features with clinical history embeddings from RoBERTa. However, these approaches rely on computationally intensive vision transformers and require costly bounding box annotations for detection.


The integration of clinical metadata into CAD systems reveals further opportunities. Zheng et al.~\cite{zheng2022multi} utilized tabular data alongside CT images to predict patient survival outcomes, while Hager et al.~\cite{hager2023best} aligned cardiac MR images with routine clinical parameters using contrastive learning. For mammography specifically, MMBCD's cross-attention mechanism between clinical history and regions of interest improved architectural distortion detection (72.3\% vs. 49\% sensitivity), yet its dependency on FocalNet-DINO for region proposals introduces annotation bottlenecks.

{\textbf{Contributions:}} The recent proposed methods and progress still do not offer an appropriate approach for deployment in real-world clinical scenarios using the full capacity of VLM models. Our work advances the clinically practical VLM models for training and inference time by: (1) A hierarchical tokenization module converting only structured metadata (age, device type, nationality, BI-RADS, density) into dense vectors compatible with convolutional features, eliminating vision-transformer overhead and utilizing stronger inductive biases; (2) multi-stage fusion blocks enabling bidirectional vision-language integration while preserving spatial relationships at native $2K$ resolution; and (3) Synthetic report generation from clinical information templates, augmenting limited text data without requiring manual annotations and extensive prior annotations. By maintaining convolutional architectures and using multi-modal tokenization, our framework provides seamless training/inference without known issues of high-resolution ViT/CLIP-based approaches while achieving superior performance on malignancy and calcification abnormality detection (AUC 0.921 vs. 0.856).


Our results demonstrate that efficient integration of vision and language features can change the performance expectations from CAD systems while maintaining scalability for real-world clinical applications.

\section{Method}
\begin{figure}[t]
    \begin{center}
    \includegraphics[width=\columnwidth]
    {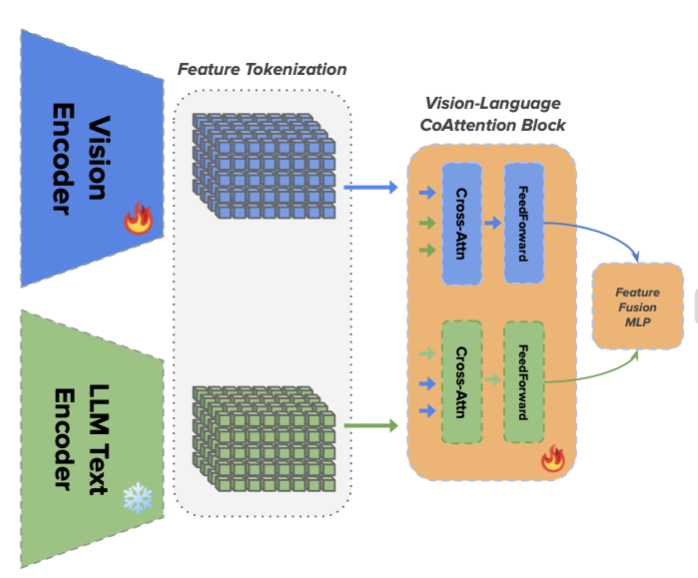}
    \vspace{-0.5cm}
    \captionsetup{font=footnotesize}
    \caption{Overview of our proposed method: a vision-language training/inference pipeline to use standard ConvNets and LLMs to integrate multi-modal information using Co-Attention mechanism and joint feature representation learning. }
    \vspace{-1cm}
    \label{fig:method}
    \end{center}
\end{figure}

The model is depicted in figure \ref{fig:method} transforms tabular data into text form and subsequently runs a VLM fine-tuning with both modalities to solve a classification task. Assuming a multi-modal dataset of paired mammography images and general patient information in the tabular form, we first transform the tabular meta-information into short text enabling the usage of VLMs. 

\noindent
\textbf{Tabular to Text} The first step is to translate the tabular data into text form, allowing us to utilize the VLM setting for efficient multi-modal learning fully. The tabular metadata (age, nationality, imaging device manufacturer and model name, institution, exam year, and breast density) is transformed into concise synthetic medical reports for each mammography image. The values of the covariates are extracted and put into a set of pre-defined sentences for fast and accurate report generation that circumvents the computational requirements of running large language models (LLMs) and prevent hallucination of smaller LLMs, which might introduce undesired noise in the pseudo reports. Missing data is handled by extracting \textit{unknown} as the value e.g. a patient of unknown age.

\noindent
\textbf{Textual Encoding} With the generated mammography reports per image, one can now fully utilize language models to extract semantic information from tabular data. The synthetic report data is fed into a language model $f_{\theta}^{text}$ obtaining token representations $T_{i}^{text}$. $T_{i}^{text}$ contains context-aware fine-grained information that might be lost in the global sequence representation given by the CLS embeddings. Another reason for the choice of the token representations is simply due to the nature of the text data set; although as close to the clinical reality as possible, they provide only limited information. Hence, token embeddings are favorable as they preserve word-specific meanings, which provide granular textual information.

\noindent
\textbf{Visual Encoding} For the image data, a vision encoder is used to extract visual information for a later fusion of the textual and visual information. The vision encoder $f_{\theta}^{vis}$ generates a feature map $F_i$, containing spatial information, indicating where features occur in an image as well as the local features at each position of the image. Researchers utilizing transformer aggregation modules with ConvNets rely on the embeddings~\cite{lee2023enhancing,yala2021toward,jain2024mmbcd}, while using the $F_i$ is not only more informative but handy for subsequent attention mechanism for later aggregation.


\noindent
\textbf{Tokenizer} In order to generate tokens from $F_i$, we initially reshape it maintaining the channel dimensions and turning the height and width into number of tokens. Since mammograms are rather large, we also use a linear projection, mapping the rearranged feature map into a more manageable number of tokens $N$. Text tokens are projected onto the same dimensionality, where the token dimensions are subject to a linear projection onto the channel dimensions of the feature map and the number of text tokens is projected onto $N$, either with adaptive average pooling in the case of a down-sampling or a linear projection in case of up-sampling.

\noindent
\textbf{Multi-Modal Fusion}
To integrate the textual information with the visual features, co-attention is utilized~\cite{hendricks2021decoupling}, which consists of two intertwined transformer blocks. Each incorporating a self-attention~\cite{vaswani2017attention} followed by a cross attention~\cite{chen2021crossvit} module and a 2-layer multi-layer perceptron (MLP). Given the token representations of both modalities, each token representation has a multi-head self-attention module applied separately, allowing the isolated features to refine its internal representation, a self-regularization. Self-attention captures long-range dependencies in the input data. The subsequent cross-attention modules allow for interaction between the modalities, such that the visual features attend to the textual features and vice versa, ensuring that both modalities become aware of each other. Hence aiding multi-modal learning. Naturally, we apply residual connections as well as a layer normalization after each attention block and the final MLP. In this sense, co-attention enables fine-grained feature interaction between textual and visual representations while also maintaining modality-specific contextual information. We apply the co-attention transformers $k$ consecutive times.

\noindent
\textbf{Classifier}
Both token representations are of high dimensionality as transformers are isomorphic by definition. To lessen the computational burden, each representation is subjected to a max-pooling. Finally, we concatenate the now pooled textual-aware visual and pooled visual-aware textual representations from the transformer and feed it to a 2-layer fusion MLP before a classification layer. The classification objective is a standard binary cross-entropy loss function.

\section{Experiment}
\subsection{Datasets}
The proposed method was developed and evaluated on two in-house mammography datasets. We evaluate the model on two different tasks: (1) malignancy classification, where the task is to separate mammograms containing biopsy-proven malignancies from all others, and (2) calcification classification, where mammograms with and without calcifications are separated. 

\noindent \textbf{BRC-dataset 1} comprises 7,454 exams (29,610 images with 4,062 are cancer cases). The dataset includes 1,511 calcification cases. The training set has 6,086 exams (3,831 cancer, 1,408 calcification) and 648 exams(221 cancer cases, 103 calcification cases) in the test set. \textbf{BRC-dataset 2} consists of 39,023 exams (144,573 images with 4,536 cancer cases). In terms of calcification, the dataset has 1123 cases. We partitioned this dataset into a training and test set containing 36,225 exams (4,038 cancer cases, 997 calcification cases) and 2,798 exams (1,173 cancer cases, 126 calcification cases), respectively. Futher, both datasets were collected from different continents.




\subsection{Implementation Details}
\textbf{Image Transformation:} Grey scale mammograms are copied 3 times and treated as RGB images with three color channels. Pixel values $<40$ in the mammograms are turned to zero, as it denotes the background~\cite{Ghosh2024MammoCLIP}. A breast region cropping is applied to isolate the breast before resizing the images to $[1520, 912]$ and then augmented by affine transformation with rotations up to 20 degrees, a minimum translation of 0.1\%, scaling factors [0.8, 1.2], and shearing by 20 degrees and elastic transformations with ($\alpha = 10, \sigma = 5$)~\cite{Ghosh2024MammoCLIP}.

\noindent
\textbf{Network Architectures:}
For the text encoder, BioClinicalBERT~\cite{alsentzer2019publicly} is used and frozen. We utilize ResNet-34~\cite{he2016deep} or EfficientNet-B5~\cite{Tan2019EfficientNet} as the vision encoder. All the vision and CLIP-based models are further initialized with our own weights from a contrastive VLM pre-training on $630,627$ annotated mammograms. The aggregation has three co-attention transformers, where the self-attention and cross-attention use four heads. The fusion MLP has $1024$ hidden and $512$ output dimensions with a Gaussian error linear unit activation.

\noindent
\textbf{Optimization:}
AdamW~\cite{loshchilov2017decoupled} optimizer is used with a learning rate of $5e$-$5$ and a weight decay of $1e$-$4$. A cosine-annealing scheduler with warm-up for $1$ epoch is used~\cite{loshchilovstochastic} as well. The training was conducted in a distributed data parallelism~\cite{li2020pytorch} setting with mixed-precision on $8$ H100 GPUs and trained for a maximum of 30 epochs, where models with a ResNet-34 encoder had a per-device mini-batch size of $96$ and EfficientNet-B5 ones $16$.

\noindent
\textbf{Baseline:}
The baseline competitors are constructed with self-attention ($8$ heads, $4$ blocks) for the vision only case, and merged-attention~\cite{dou2022empirical} and cross-attention with $4$ and $3$ blocks following~\cite{dou2022empirical}. The naive MLP aggregator concatenates the image and text embeddings before directly passing through the fusion MLP.
\vspace{-0.1cm}
\subsection{Results}
Table \ref{tab:main} shows the performance of our method compared to the baseline of image-only models as well as merged-attention and cross-attention aggregators to incorporate text guidance into the classification. Our method outperforms image-only models, while co-attention emerges as the Pareto optimal aggregator across two different datasets for calcification and malignancy classification. Malignancy classification shows an improvement compared to a common classification model of $6$ \% points for ResNet and $3$ \% for EfficientNet on BRC1 while the improvements on BRC2 were $5$ and $2.4$ \% for ResNet and EfficientNet, respectively. At the calcification classification on BRC1, we present AUC gains of $3.3$ and $2.1$ \% for both backbones. On BRC2 our model gains $6.3$ and $3.5$ \% AUC. Additionally, our method also outperforms a transformer model on top of a vision backbone with steady improvements of at least $2$\% points in AUC across all setting. For ResNet backbones, our models have shown that more data improves the performance, as combining the two datasets during training leads to a consistent improvements in AUC. 

Overall, we have shown that aggregation via co-attention on top of simple ConvNets is an easy and efficient way to improve clinical predictions with the only addition of limited tabular data.

\begin{table*}[!ht]
    \centering
   
    \resizebox{0.48\textwidth}{!}{
    \begin{tabular}{l|l|l|c|c|c|c}
    \toprule
    \multirow{2}{*}{Method} &
        \multirow{2}{*}{Aggregation} &
        \multirow{2}{*}{Encoder} &
        \multicolumn{2}{c|}{Malignancy} &
        \multicolumn{2}{c}{Calcification} \\
    \cline{4-7}
     &  &  & BRC1 & BRC2 & BRC1 & BRC2 \\
    \midrule
    
     Vision-Only & None & RN34 & 0.8594 & 0.8352 & 0.9108 & 0.8575 \\

     Vision-Only & Self-Attention & RN34 & 0.8940 & 0.8686 & 0.9223 & 0.8798 \\

     Text Guided & Naive MLP & RN34 & 0.8816 & 0.8687 & 0.9198 & 0.8806 \\

     Text Guided & Merged-Attention & RN34 & 0.9148 & 0.8837 & 0.9317 & 0.9105 \\

     Text Guided & Cross-Attention & RN34 & 0.9187 & 0.8815 & 0.9448 & 0.9143 \\

     Text Guided & Co-Attention & RN34 & 0.9147 & 0.8864 & 0.9358 & 0.9205 \\

     Text Guided* & Merged-Attention & RN34 & 0.9311 & 0.8867 & 0.9344 & 0.9190\\

     Text Guided* & Cross-Attention & RN34 & 0.9295 & 0.8866 & 0.9450 & 0.9220 \\

     Text Guided* & Co-Attention & RN34 & \textbf{0.9320} & \textbf{0.8870} & \textbf{0.9452} & \textbf{0.9267} \\

     \hline

     Vision-Only & None & ENB5 & 0.8992 & 0.8624 & 0.9219 & 0.8822 \\

     Vision-Only & Self-Attention & ENB5 & 0.9076 & 0.8748 & 0.9279 & 0.8871 \\

     Text Guided & Naive MLP & ENB5 & 0.9083 & 0.8760 & 0.9308 & 0.8926 \\

     Text Guided & Merged-Attention & ENB5 & \textbf{0.9280} & 0.8823 & 0.9380 & 0.9125 \\

     Text Guided & Cross-Attention & ENB5 & 0.9229 & 0.8855 & 0.9415 & 0.9144 \\

     Text Guided & Co-Attention & ENB5 & 0.9247 & \textbf{0.8864} & \textbf{0.9434} & \textbf{0.9186} \\
    
    \bottomrule
    \end{tabular}
    }
    \captionsetup{font=footnotesize}
    \caption{Performance of our model on Malignancy and Calcification classification, evaluated with the AUC. Text-guided (ours) models are compared with various multi-modal transformer aggregation techniques as well as an MLP aggregation (naive MLP) and vision-only models. The baseline is an image encoder with a classification head with and without transformer aggregation. * marks models trained on both BRC1 and 2.}
    \label{tab:main}
\end{table*}

\subsection{Ablation Experiments}

\begin{figure}[ht]
    \vspace{-5mm}
    \centering
    \scalebox{0.75}{  
        \begin{minipage}{0.65\linewidth}
            \centering
            \includegraphics[width=\linewidth]{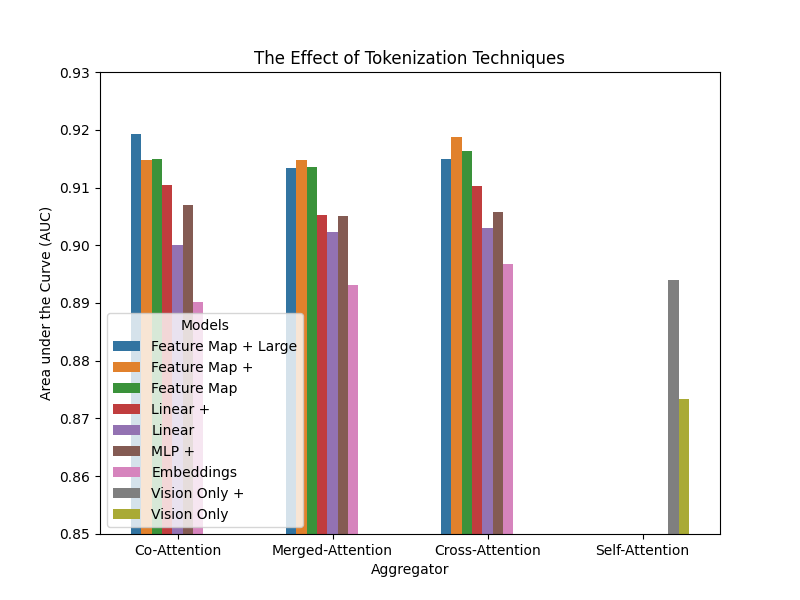}
            \subcaption{The effect of various tokenizers.}
        \end{minipage}%
        \hspace{0pt}  
        \begin{minipage}{0.65\linewidth}
            \centering
            \includegraphics[width=\linewidth]{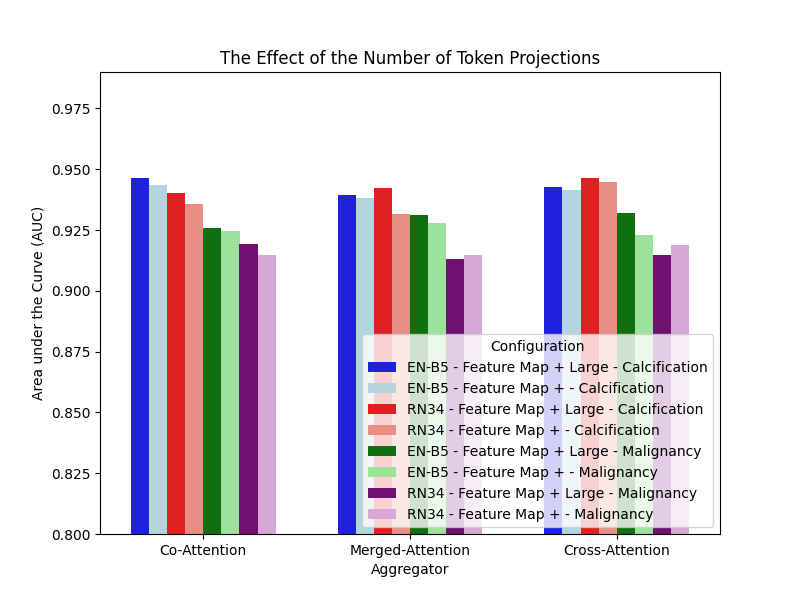}
            \subcaption{The effect of the number of tokens.}
        \end{minipage}
    }
    \captionsetup{font=footnotesize}
    \caption{Effectiveness of tokenizers (a) and the number of tokens (b) on the malignancy classification performance on BRC1. (a) contains the malignancy classification performance of a ResNet34 Vision Backbone for various transformer aggregation techniques. (b) relates the number of tokens to the classification performance for malignancy and calcification. + indicated a model with max pooling, while Large marks models with 512 tokens. All other models work on 256 tokens. The classification performance is evaluated by AUC.}
    \label{fig:token_shit}
    \vspace{-5mm}
\end{figure}

Figure \ref{fig:token_shit} displays the effect of tokenizers (a) and the number of tokens (b) on the classification performance. Sub-figure (a) supports using the feature map as single tokens generated from embeddings hinder the classification accuracy. Further, we can show that linear and MLP tokenizers, generating tokens from embeddings directly, are less predictive than any feature map setting for the same amount of tokens produced. Max-pooling was shown to be beneficial to the classifier. The number of tokens can be shown to aid the classifier, as all models perform better with $512$ tokens compared to $256$, although the improvements were marginal. This holds for both ConvNets with all three aggregators as can be depicted from subfigure (b) in figure \ref{fig:token_shit}.

\begin{table}[!ht]
    \vspace{-5mm}
    \centering
    \footnotesize 
    \resizebox{0.80\columnwidth}{!}{
    \begin{tabular}{l|l|c|c}
    \toprule
    \multirow{2}{*}{Method} &
        \multirow{2}{*}{Encoder} &
        {RSNA} &
        {VinDR} \\
    \cline{3-4}
     &  & Malignancy & Calcification\\
    \midrule
    
     Supervised & EN-B5 & 0.7271 & 0.9654\\
     
     Mammo-CLIP~\cite{Ghosh2024MammoCLIP} & EN-B5 & 0.7257 &  0.9746\\
     
     CLIP~\cite{radford2021learning} & EN-B5 & 0.7659 & 0.9768 \\

     MV-CLIP & EN-B5  &  0.7620 & 0.9787\\

     MaMa-CLIP~\cite{du2024multi} & ViT-B-14 &  0.7300 & -- \\

     MGCA~\cite{wang2022multi} & ViT-B-14 & 0.6870 & --\\

     MM-MIL~\cite{wang2023using} & ViT-B-14 & 0.6500 & --\\

     Ours* & EN-B5 & 0.7837 & \textbf{0.9806} \\

     Ours** & EN-B5 & \textbf{0.7928} & 0.9760 \\

    \bottomrule
    \end{tabular}
    }
    \captionsetup{font=footnotesize}
    \caption{Malignancy and calcification classification performance of the vision backbone extracted from our model on the RSNA Mammo\cite{rsnamm} and VinDr Mammo\cite{vindr} datasets, evaluated by AUC. * denotes training on BRC 1, while ** indicates models trained on BRC 2. Each model is trained on the respective classification task. The CLIP is pre-trained using our own Mammography dataset with a higher resolution, while MV-CLIP is pre-trained in the same manner with multi-view alignment.}
    \label{tab:public}
     \vspace{-5mm}
\end{table}

We also evaluated our model on the public benchmark datasets VinDr~\cite{vindr} and RSNA Mammo~\cite{rsnamm}. The evaluation is conducted by with a fine-tuned vision backbone trained on either malignancy or calcification classification. Table \ref{tab:public} shows these results. The vision backbone trained with our method shows improvements in malignancy classification on RSNA mammo, irrespective of whether it was trained on BRC1 or BRC2. For RSNA, we presented the state-of-the-art performances and improvements of almost $3$\% AUC compared to a customized CLIP pre-trained model. Calcification classification also shows slight improvements compared to the baseline, although the $0.2$ \% AUC gains are low in number, which might indicate saturation of the VinDr dataset. Overall, the results support our method, as integrating superficial metadata seems to also aid in finding better pre-trained vision backbones.

\section{Discussion}
This study introduces a practical clinical-level vision-language model for breast cancer CAD that effectively detects malignancy and calcification across diverse datasets. Our scalable framework seamlessly integrates mammography images with existing clinical text information as extra context. The implementation of auxiliary fine-grained feature map tokenization with multi-modal aggregation significantly enhances the detection of minuscule imaging variations, particularly benefiting calcification classification while maintaining computational efficiency.

{
    \small
    \bibliographystyle{ieeenat_fullname}
    \bibliography{references}

\begin{thebibliography}{25}
\providecommand{\natexlab}[1]{#1}
\providecommand{\url}[1]{\texttt{#1}}
\expandafter\ifx\csname urlstyle\endcsname\relax
  \providecommand{\doi}[1]{doi: #1}\else
  \providecommand{\doi}{doi: \begingroup \urlstyle{rm}\Url}\fi

\bibitem[Alsentzer et~al.(2019)Alsentzer, Murphy, Boag, Weng, Jin, Naumann, and McDermott]{alsentzer2019publicly}
Emily Alsentzer, John~R Murphy, Willie Boag, Wei-Hung Weng, Di Jin, Tristan Naumann, and Matthew McDermott.
\newblock Publicly available clinical bert embeddings.
\newblock \emph{arXiv preprint arXiv:1904.03323}, 2019.

\bibitem[Carr et~al.()Carr, Kitamura, Kalpathy-Cramer, Mongan, Andriole, Vazirabad, Riopel, Ball, and Dane]{rsnamm}
Chris Carr, Felipe Kitamura, J Kalpathy-Cramer, J Mongan, K Andriole, M Vazirabad, M Riopel, R Ball, and S Dane.
\newblock Rsna screening mammography breast cancer detection. 2022.

\bibitem[Chen et~al.(2021)Chen, Fan, and Panda]{chen2021crossvit}
Chun-Fu~Richard Chen, Quanfu Fan, and Rameswar Panda.
\newblock Crossvit: Cross-attention multi-scale vision transformer for image classification.
\newblock In \emph{Proceedings of the IEEE/CVF international conference on computer vision}, pages 357--366, 2021.

\bibitem[Dou et~al.(2022)Dou, Xu, Gan, Wang, Wang, Wang, Zhu, Zhang, Yuan, Peng, et~al.]{dou2022empirical}
Zi-Yi Dou, Yichong Xu, Zhe Gan, Jianfeng Wang, Shuohang Wang, Lijuan Wang, Chenguang Zhu, Pengchuan Zhang, Lu Yuan, Nanyun Peng, et~al.
\newblock An empirical study of training end-to-end vision-and-language transformers.
\newblock In \emph{Proceedings of the IEEE/CVF Conference on Computer Vision and Pattern Recognition}, pages 18166--18176, 2022.

\bibitem[Du et~al.(2024)Du, Onofrey, and Dvornek]{du2024multi}
Yuexi Du, John Onofrey, and Nicha~C Dvornek.
\newblock Multi-view and multi-scale alignment for contrastive language-image pre-training in mammography.
\newblock \emph{arXiv preprint arXiv:2409.18119}, 2024.

\bibitem[Elmore et~al.(2005)Elmore, Armstrong, Lehman, and Fletcher]{elmore2005screening}
Joann~G Elmore, Katrina Armstrong, Constance~D Lehman, and Suzanne~W Fletcher.
\newblock Screening for breast cancer.
\newblock \emph{Jama}, 293\penalty0 (10):\penalty0 1245--1256, 2005.

\bibitem[Evans et~al.(2013)Evans, Birdwell, and Wolfe]{evans2013if}
Karla~K Evans, Robyn~L Birdwell, and Jeremy~M Wolfe.
\newblock If you don’t find it often, you often don’t find it: why some cancers are missed in breast cancer screening.
\newblock \emph{PloS one}, 8\penalty0 (5):\penalty0 e64366, 2013.

\bibitem[Ghosh et~al.(2024)Ghosh, Chen, Li, and Xie]{Ghosh2024MammoCLIP}
Arindam Ghosh, Xuxin Chen, Yuxuan Li, and Weidi Xie.
\newblock {Mammo-CLIP}: A vision language foundation model to enhance data efficiency and robustness in mammography.
\newblock \emph{arXiv preprint arXiv:2404.15946}, 2024.

\bibitem[Hager et~al.(2023)Hager, Menten, and Rueckert]{hager2023best}
Paul Hager, Martin~J Menten, and Daniel Rueckert.
\newblock Best of both worlds: Multimodal contrastive learning with tabular and imaging data.
\newblock In \emph{Proceedings of the IEEE/CVF Conference on Computer Vision and Pattern Recognition}, pages 23924--23935, 2023.

\bibitem[He et~al.(2016)He, Zhang, Ren, and Sun]{he2016deep}
Kaiming He, Xiangyu Zhang, Shaoqing Ren, and Jian Sun.
\newblock Deep residual learning for image recognition.
\newblock In \emph{Proceedings of the IEEE conference on computer vision and pattern recognition}, pages 770--778, 2016.

\bibitem[Hendricks et~al.(2021)Hendricks, Mellor, Schneider, Alayrac, and Nematzadeh]{hendricks2021decoupling}
Lisa~Anne Hendricks, John Mellor, Rosalia Schneider, Jean-Baptiste Alayrac, and Aida Nematzadeh.
\newblock Decoupling the role of data, attention, and losses in multimodal transformers.
\newblock \emph{Transactions of the Association for Computational Linguistics}, 9:\penalty0 570--585, 2021.

\bibitem[Jain et~al.(2024)Jain, Bansal, Rangarajan, and Arora]{jain2024mmbcd}
Kshitiz Jain, Aditya Bansal, Krithika Rangarajan, and Chetan Arora.
\newblock Mmbcd: Multimodal breast cancer detection from mammograms with clinical history.
\newblock In \emph{International Conference on Medical Image Computing and Computer-Assisted Intervention}, pages 144--154. Springer, 2024.

\bibitem[Lee et~al.(2023)Lee, Kim, Park, Kim, Kim, and Kooi]{lee2023enhancing}
Hyeonsoo Lee, Junha Kim, Eunkyung Park, Minjeong Kim, Taesoo Kim, and Thijs Kooi.
\newblock Enhancing breast cancer risk prediction by incorporating prior images.
\newblock In \emph{International Conference on Medical Image Computing and Computer-Assisted Intervention}, pages 389--398. Springer, 2023.

\bibitem[Lehman et~al.(2015)Lehman, Wellman, Buist, Kerlikowske, Tosteson, Miglioretti, Consortium, et~al.]{lehman2015diagnostic}
Constance~D Lehman, Robert~D Wellman, Diana~SM Buist, Karla Kerlikowske, Anna~NA Tosteson, Diana~L Miglioretti, Breast Cancer~Surveillance Consortium, et~al.
\newblock Diagnostic accuracy of digital screening mammography with and without computer-aided detection.
\newblock \emph{JAMA internal medicine}, 175\penalty0 (11):\penalty0 1828--1837, 2015.

\bibitem[Li et~al.(2020)Li, Zhao, Varma, Salpekar, Noordhuis, Li, Paszke, Smith, Vaughan, Damania, et~al.]{li2020pytorch}
Shen Li, Yanli Zhao, Rohan Varma, Omkar Salpekar, Pieter Noordhuis, Teng Li, Adam Paszke, Jeff Smith, Brian Vaughan, Pritam Damania, et~al.
\newblock Pytorch distributed: Experiences on accelerating data parallel training.
\newblock \emph{arXiv preprint arXiv:2006.15704}, 2020.

\bibitem[Loshchilov(2017)]{loshchilov2017decoupled}
I Loshchilov.
\newblock Decoupled weight decay regularization.
\newblock \emph{arXiv preprint arXiv:1711.05101}, 2017.

\bibitem[Loshchilov and Hutter()]{loshchilovstochastic}
I Loshchilov and F Hutter.
\newblock Stochastic gradient descent with warm restarts.
\newblock In \emph{Proceedings of the 5th International Conference on Learning Representations}, pages 1--16.

\bibitem[Nguyen et~al.(2023)Nguyen, Nguyen, Pham, Lam, Le, Dao, and Vu]{vindr}
Hieu~T Nguyen, Ha~Q Nguyen, Hieu~H Pham, Khanh Lam, Linh~T Le, Minh Dao, and Van Vu.
\newblock Vindr-mammo: A large-scale benchmark dataset for computer-aided diagnosis in full-field digital mammography.
\newblock \emph{Scientific Data}, 10\penalty0 (1):\penalty0 277, 2023.

\bibitem[Radford et~al.(2021)Radford, Kim, Hallacy, Ramesh, Goh, Agarwal, Sastry, Askell, Mishkin, Clark, et~al.]{radford2021learning}
Alec Radford, Jong~Wook Kim, Chris Hallacy, Aditya Ramesh, Gabriel Goh, Sandhini Agarwal, Girish Sastry, Amanda Askell, Pamela Mishkin, Jack Clark, et~al.
\newblock Learning transferable visual models from natural language supervision.
\newblock In \emph{International conference on machine learning}, pages 8748--8763. PmLR, 2021.

\bibitem[Tan and Le(2019)]{Tan2019EfficientNet}
Mingxing Tan and Quoc~V. Le.
\newblock {EfficientNet}: Rethinking model scaling for convolutional neural networks.
\newblock In \emph{{Proceedings of the International Conference on Machine Learning (ICML)}}, pages 6105--6114, 2019.

\bibitem[Vaswani et~al.(2017)Vaswani, Shazeer, Parmar, Uszkoreit, Jones, Gomez, Kaiser, and Polosukhin]{vaswani2017attention}
Ashish Vaswani, Noam Shazeer, Niki Parmar, Jakob Uszkoreit, Llion Jones, Aidan~N Gomez, Łukasz Kaiser, and Illia Polosukhin.
\newblock Attention is all you need.
\newblock \emph{Advances in Neural Information Processing Systems}, 30, 2017.

\bibitem[Wang et~al.(2022)Wang, Zhou, Wang, Vardhanabhuti, and Yu]{wang2022multi}
Fuying Wang, Yuyin Zhou, Shujun Wang, Varut Vardhanabhuti, and Lequan Yu.
\newblock Multi-granularity cross-modal alignment for generalized medical visual representation learning.
\newblock \emph{Advances in Neural Information Processing Systems}, 35:\penalty0 33536--33549, 2022.

\bibitem[Wang et~al.(2023)Wang, Wells, Berkowitz, Horng, and Golland]{wang2023using}
Peiqi Wang, William~M Wells, Seth Berkowitz, Steven Horng, and Polina Golland.
\newblock Using multiple instance learning to build multimodal representations.
\newblock In \emph{International Conference on Information Processing in Medical Imaging}, pages 457--470. Springer, 2023.

\bibitem[Yala et~al.(2021)Yala, Mikhael, Strand, Lin, Smith, Wan, Lamb, Hughes, Lehman, and Barzilay]{yala2021toward}
Adam Yala, Peter~G Mikhael, Fredrik Strand, Gigin Lin, Kevin Smith, Yung-Liang Wan, Leslie Lamb, Kevin Hughes, Constance Lehman, and Regina Barzilay.
\newblock Toward robust mammography-based models for breast cancer risk.
\newblock \emph{Science Translational Medicine}, 13\penalty0 (578):\penalty0 eaba4373, 2021.

\bibitem[Zheng et~al.(2022)Zheng, Lin, Zhou, Peng, Xiao, Zu, Jiao, and Wang]{zheng2022multi}
Hanci Zheng, Zongying Lin, Qizheng Zhou, Xingchen Peng, Jianghong Xiao, Chen Zu, Zhengyang Jiao, and Yan Wang.
\newblock Multi-transsp: Multimodal transformer for survival prediction of nasopharyngeal carcinoma patients.
\newblock In \emph{International Conference on Medical Image Computing and Computer-Assisted Intervention}, pages 234--243. Springer, 2022.

\end{thebibliography}
}

\end{document}